\RequirePackage{amsmath}
\documentclass[runningheads]{llncs}
\usepackage{graphicx}
\usepackage{mathtools}
\usepackage{multirow}
\usepackage{graphicx}\usepackage{amsmath}
\usepackage{amssymb}
\usepackage[tight,footnotesize]{subfigure}
\usepackage{epstopdf}
\usepackage{makecell}
\usepackage[table]{xcolor}
\usepackage[flushleft]{threeparttable}
\usepackage{color}
\usepackage{comment}
\graphicspath{{.}{Figures/}}
\usepackage{pifont}
\newcommand{\cmark}{\ding{52}}%
\AppendGraphicsExtensions{.tif}

\definecolor{mygrey}{HTML}{d9d9d9}

\begin{document}

\title{Online Dynamic Reliability Evaluation of Wind Turbines based on Drone-assisted Monitoring}

\titlerunning{Online Dynamic Reliability Evaluation of Wind Turbines}

\author{Sohag Kabir\inst{1} \and
Koorosh Aslansefat\inst{2} \and
Prosanta Gope\inst{3} \and 
Felician Campean\inst{1} \and 
Yiannis Papadopoulos\inst{2} }

\authorrunning{S. Kabir et al.}

\institute{
    \inst{}
        Faculty of Engineering and Informatics, University of Bradford, Bradford, UK \\
        \email{\{s.kabir2, f.campean\}@bradford.ac.uk}
    \and
    \inst{}
        Department of Computer Science and Technology, University of Hull, Hull, UK  \\
        \email{\{k.aslansefat, c.walker-2018, c.rothon-2017, y.i.papadopoulos\}@hull.ac.uk}
    \and
    \inst{}
        Department of Computer Science, University of Sheffield,Sheffield, UK \\
        \email{p.gope@sheffield.ac.uk}
   }

\maketitle             

\begin{abstract}
The offshore wind energy is increasingly becoming an attractive source of energy due to having lower environmental impact. Effective operation and maintenance that ensures the maximum availability of the energy generation process using offshore facilities and minimal production cost are two key factors to improve the competitiveness of this energy source over other traditional sources of energy. Condition monitoring systems are widely used for health management of offshore wind farms to have improved operation and maintenance. Reliability of the wind farms are increasingly being evaluated to aid in the maintenance process and thereby to  improve the availability of the farms. However, much of the reliability analysis is performed offline based on statistical data.  In this article, we propose a drone-assisted monitoring based method for online reliability evaluation of wind turbines. A blade system of a wind turbine is used as an illustrative example to demonstrate the proposed approach.

\end{abstract}

\section{Introduction}
\label{sec1}

To have a sustainable and clean planet, renewable energies play a vital role.  Among the renewable energy sources, offshore wind energy is a promising one. It is estimated that by 2030, up to 7.7\% of Europe's overall electricity demand will be fulfilled by offshore wind energy through the installation of wind farms with 66 Gigawatt production capacity \cite{corbetta2015wind}.  The only way to achieve this and to make this type of energy widespread, it should be cost-competitive with traditional energy sources. Thus, the Levelized Cost of Energy (LCoE) generated by offshore wind farms, currently valued at 80–100 €/MWh\cite{scheu2017influence}, should be reduced down to an acceptable level. The long-term forecast of UK government has envisaged that by 2050, the LCoE of offshore wind energy to be reduced to the level close to that of today's LCoE of onshore wind energy, i.e., approximately 60 €/MWh \cite{TINA2012}.  

Turbines efficiency and production rely on operation and maintenance procedure that consumes the LCoE up to 35 percent \cite{martin2016sensitivity}.
The overall inspection and repair time for an offshore wind farm can randomly prolong due to inaccessibility issues caused by safety restrictions, inappropriate weather, etc.  
Due to inaccessibility problems, an onshore wind turbine with 97 percent availability can have 76 percent availability when it is located in about 15 km offshore \cite{dowell2013analysis}. Therefore, reliability and availability are critical factors for cost-efficient production of energy using offshore wind systems.  

To improve the performance of wind turbines different types of maintenance tasks are performed. During maintenance activities, failed or degraded parts of the wind turbines are either replaced or repaired. Among the different types of maintenance strategies, corrective maintenance is performed after a component fails, i.e., on-demand maintenance is performed. The sudden need for such maintenance without any prior indication or preparation coupled with other associated factors such as asset availability, weather conditions can lead to prolonged downtime, thus making this type maintenance very costly. To avoid such unexpected sudden maintenance, preventive maintenance strategies, where scheduled maintenance are performed in the predefined interval, are used to perform the repair and/or replacement tasks before a failure occurs. To optimize the maintenance schedule, the health of the components of the wind turbine systems are monitored using condition monitoring systems (CMS). Based on the monitoring knowledge, the health status of the components are determined and maintenance tasks for components are decided based on their criticality and current health condition. 

As the reliability of wind turbines plays a crucial role in their performance and maintenance decision, numerous efforts have been made for reliability analysis of wind turbines. Different reliability analysis models such as fault trees \cite{kang2019fault}, Markov chains \cite{adumene2020markovian}, Bayesian networks (BN) \cite{adedipe2020bayesian}, 
etc. have been utilised to model the failure behaviour of wind turbines. Subsequently, statistical failure data of wind turbines' components are used in these models to evaluate the wind turbine reliability. In most of the cases, as the practical operational scenarios or profiles of the components are not taken into account, such reliability evaluation fails to consider the practical scenario properly. Therefore, changes in reliability under different operational conditions should be determined for effective preventive maintenance and to improve system operation. As the factors affecting reliability changes continuously over time, real-time/online reliability evaluation is an ideal solution for capturing effects of such changes in reliability. Some of the above-mentioned techniques have been adapted to perform online reliability analysis where data monitored and recorded by sensors and instruments are utilised to update real-time reliability. 

As blades are a major component of wind turbines, up to 25 percent of the production of a turbine depends on its healthy blades \cite{herring2019increasing}. Hence, failure of blades could lead to unscheduled downtime and substantial monetary loss.  Variable wind loads and other environmental conditions including dust, lightening, and icy weather make blades a fragile component of wind turbines. For this reason, reliability analysis and health monitoring of blades received significant interest from industry and academia. Rope-based inspection by human and telescope observation are traditionally used for blades’ health monitoring and defects detection on the surface of the blades. These approaches are time-consuming, expensive and they have low detection efficiency and high-risk factors. Different sensors such as acoustic emission (AE) sensors, vibration sensors, ultrasonic sensors and strain sensors have been used for the monitoring of blades\cite{marquez2012condition}.
 Other more advanced sensors such as fibre Bragg grating sensors \cite{lee2015transformation} and scanning laser Doppler vibrometer sensors \cite{ozbek2013challenges} have also been utilised for wind turbines’ blades health monitoring. Although sensor-based defect detection of wind turbine blades is shown to be useful, further visual inspections are still needed to obtain the details of the defects.  

In recent decades, the utilisation of unmanned aerial vehicles (UAVs), e.g. drones, for remote monitoring has gained popularity.  In addition to monitoring civil structures such as roads and bridges, UAVs are recently being commercially used to inspect the surface conditions of wind turbines’ components,   especially the blades of wind turbines. 
Although drones are being used for structural health monitoring of wind turbines, especially for blades, the monitoring knowledge is rarely used for online reliability evaluation of wind turbine systems. In this article, we look at this issue and proposed a drone-based wind turbine monitoring architecture which aims at utilising drone-based monitoring knowledge for online evaluation of reliability.

\section{Related Works}
\label{sec2}

\subsection{Wind Turbine Maintenance}
\label{sec2.3}
Effective maintenance of wind turbines plays a key role in ensuring the cost-effective and successful operation of wind farms. Among different maintenance strategies, corrective and preventive maintenance are two widely used maintenance strategies. Under the corrective strategy, unscheduled maintenance tasks are performed when failures or faults are detected in wind energy systems during an inspection. This type of maintenance is the most expensive one both with regards to availability and cost because unavailability of spare parts may lead to the purchase of expensive parts on-demand and unfavourable conditions to perform the maintenance tasks can lead to prolonged downtime causing unavailability. On the other hand, through scheduled maintenance, the preventive strategy aimed at replacing and/or repairing components before failure occurs. While this type of maintenance performed at regular interval irrespective of the conditions of the parts will lead to improved availability, this may lead to unnecessary maintenance resulting in increased maintenance cost. To utilise the knowledge of the lifespan of components, alternative approaches such as condition-based maintenance (CBM) and reliability centred maintenance (RCM) are used \cite{tracht2013failure}. Under CBM, specific components of the wind turbine systems are continuously monitored during operation to determine the health of the components and maintenance decisions are made based on this monitoring knowledge before the components fail. In RCM, the reliability of different parts of wind turbines is estimated and compared with predefined threshold values. If the reliability of certain parts falls below the respective value of that part then the critical components that contributed the most, e.g., that degraded the most, to the reduction in reliability are identified. Afterwards, maintenance activities are planned to restore the functionality of the affected components, hence achieving the required level of reliability.     
For condition monitoring of wind farms, different condition monitoring systems (CMS) such as component-specific systems (e.g. for bearings or gearboxes) or manufacturer dependent systems (e.g. for Nordex or GE) are used within the wind turbine systems to provide component-specific information to the operators \cite{tracht2013failure}. 
 CMSs continuously monitor different parts of the system and regularly reports the monitoring information to a central data acquisition server, for example, to a supervisory control and data acquisition (SCADA) system. 
  Parameters that are usually monitored include, but are not limited to:  bearing or gearbox temperature, oil, vibrations, wind speed and directions, power output. By processing the data available in the SCADA system, it is possible to predict the component failure in advance. In \cite{kusiak2010data}, it was shown that failures of components can be predicted  5 to 60 minutes in advance. Tracht \textit{et al.} \cite{tracht2013failure} proposed a method to process monitoring data available in the SCADA system to predict the failure probability of components in wind turbine systems. This type of online predicted probabilities can be used in reliability models to obtain online reliability of systems.

Note that using traditional sensor-based monitoring system, it is not possible to monitor all parts of the wind turbines. For instance, during operation, it is not possible to inspect or monitor different conditions that can cause the failure of wind turbines' blades. As mentioned before, in many cases, on top of sensor-based monitoring, it is necessary to have manual inspections by the human to determine the degradation level of wind turbine components. To alleviate the limitations of human-based manual inspection, UAVs are increasingly being used in the offshore wind industry. The following section provides a brief overview of drone-based monitoring of offshore wind industry.

\subsection{Drone-based Monitoring}
\label{sec2.2}
In the offshore wind industry, drones have been used in different ways such as for remote inspection, facilitating maintenance activities etc. Thermography and visual inspections are two commonly used ways for UAV-based monitoring. The cost-benefit evaluation of the UAV-based or drone-based blade inspection is an essential step before using the technology in the industry and \cite{kapoor2018unmanned} has analysed the cost-benefit of UAV-based structural inspection of the wind turbines. The paper has addressed different comparisons and quantitative cost analysis between conventional wind turbine structural inspection and the UAV-based one. The report shows a significant reduction in the cost and operational time in using UAV-based inspection.

Regarding the drone-based inspection, in the experimental study shown in \cite{franko2020design}, multicopters with vision and LiDAR sensors have been used for inspection of turbine's blades to guide the climbing robots while performing maintenance of the blades of wind turbines.  A data-driven framework was proposed by Wang and Zhang \cite{wang2017automatic} for automatic crack detection on the surface of wind turbine blades. For the detection and classification of the cracks, they utilised a set of models such as the LogitBoost, Decision Tree, and Support Vector Machine. Peng and Liu \cite{peng2018detection} utilises the images taken by UAVs to propose an analytical crack detection method for wind turbine blades. At first, they performed some pre-processing of the images taken by drones to improve their quality. 
Once the noises were removed, an image enhancement algorithm was used to improve the quality of the images. Finally, grey-scale versions of the images were processed to identify and classify the cracks on the blade surface. 

Regarding the processing of drone-based inspection images, the application of deep learning for blade abnormality detection for offshore wind turbines can be an interesting subject. For these algorithms, the important issue is to be fed by tailored and high quality trusted dataset. An open-source dataset of drone-based inspection images has been provided by \cite{shihavuddin2018dtu}.
A new method for automating the data labelling for drone-based inspection images has been proposed by \cite{shihavuddin2019wind}. The paper has used different deep neural networks architectures such as Inception-V2, ResNet-50, ResNet-101 and ResNet-101. 
Reference \cite{reddy2019detection} has focused on drone-based inspection images and proposed a method for structural health monitoring of wind turbine blades based on Convolutional Neural Network (CNN) classifier. In this paper, the image annotation has been done manually and there were two types of performed binary classification (faulty and normal) with the accuracy above 94 percent, and multi-class classification with nine classes (Sky, Nature and blade, Nature, Lightening damage, Mechanical damage, Tip open, Erosion, Side-erosion and Cracks) and the accuracy above 90 percent. 
Regarding the blade inspection of offshore wind farms, a formulated and optimized route planning of UAVs has been proposed by \cite{chung2020placement} in which effect of wind speed, flying range, and flying speed of UAV has been considered. The effectiveness of the proposed method has been simulated and evaluated with  Walney wind farm recorded data.


\subsection{Reliability and Availability Assessment of Wind Turbines}
\label{ReliabilityandAvailability}
Reliability and availability are two important non-functional properties that are increasingly assessed to improve the operation and maintenance of offshore wind farms. Different research efforts have been made for reliability and availability analysis in the wind industry. Tavner \textit{et al.} \cite{tavner2007reliability} utilised 10 years (1994 to 2004) operational data extracted from Windstats (www.windstats.com) for reliability analysis of Danish and German wind turbines. In \cite{faulstich2011wind}, the importance of downtime in the offshore deployment of wind turbines and how reliability and availability are linked with downtime were discussed.  
A classification of the existing approaches for risk and reliability analysis of wind turbines including qualitative, semi-quantitative and quantitative methods has been provided by \cite{leimeister2018review}. Alhmoud et al. \cite{alhmoud2018review} have discussed the state-of-the-art on the reliability modelling of the wind turbines and addressed reliability modelling techniques, life distributions, reliability testing methodologies, reliability design and severity classifications. Igwemezie et al. \cite{igwemezie2019current} have provided the current trend of the offshore wind energies and the requirements. The paper also studied the reliability improvement of the structures in offshore wind turbines. The 127 possible failures on the offshore wind turbine and their combinations have been addressed by \cite{kang2019fault}. The paper has used the Fault Tree Analysis (FTA) method for reliability evaluation of floating offshore wind turbines and highlighted the most probable bottlenecks in offshore wind turbine through importance measure. Considering online monitoring data of a system and updating the reliability models is one of hot topics in the filed of reliability and safety engineering. For example, \cite{cai2015real} has used Dynamic BN for root cause diagnosis and online reliability evaluation of subsea blowout preventer. Kabir et al. \cite{kabir2019conceptual} have introduced the concept of complex basic events in Fault Tree with the ability of updating the basic events probabilities based on online monitoring data. 
Online monitoring data can be also used for safety model repair. Gheraibia et al. \cite{gheraibia2019safety+} has proposed a Support Vector Machine (SVM)-based approach to use online monitoring data and suggest a correction in reliability models like fault tree.
\section{Proposed Approach}
\label{sec4}
\begin{figure*}[!thpb]
      \centering
			\includegraphics[scale=0.33]{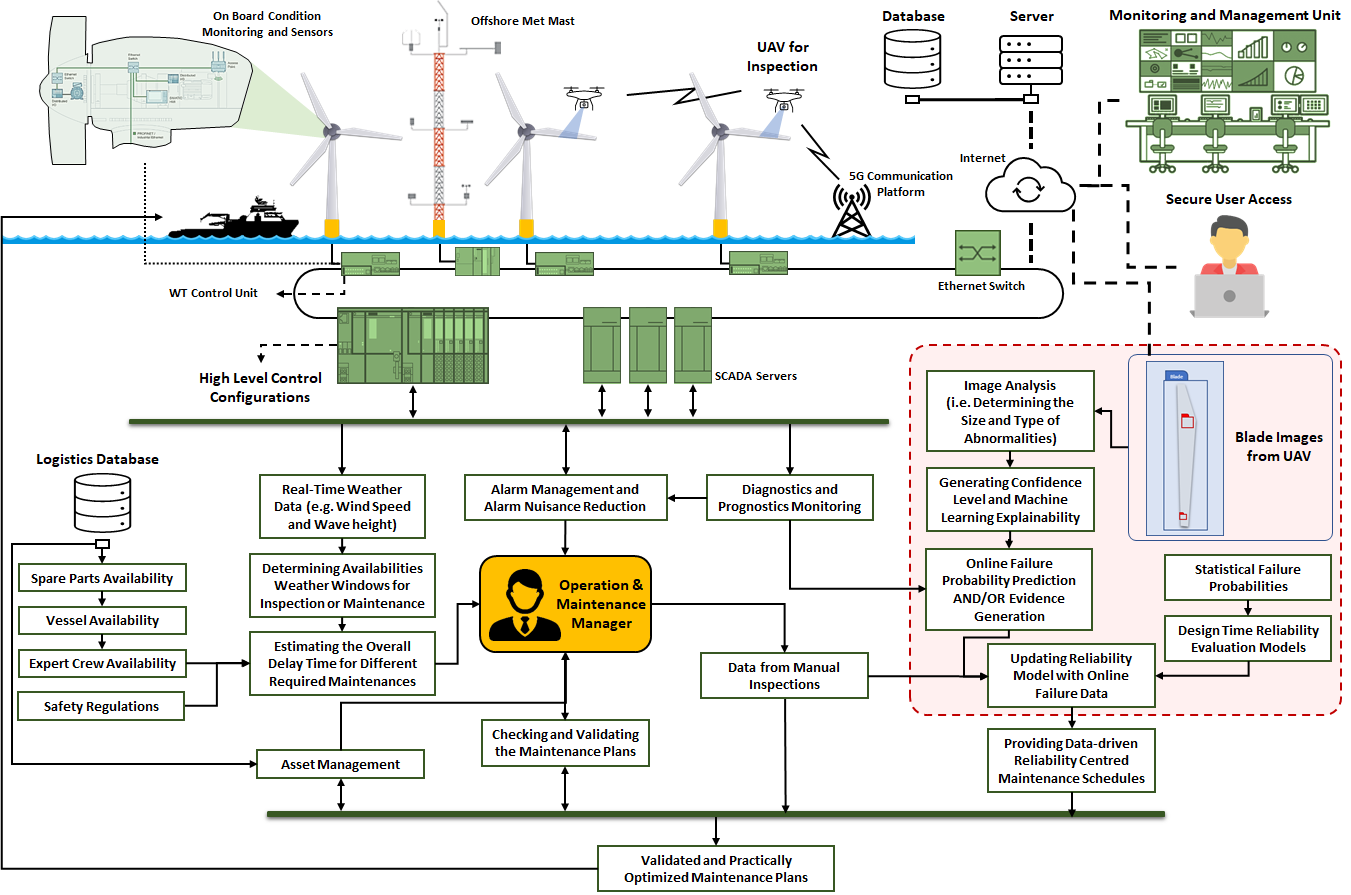}
     \caption{The structure of the complete framework}
      \label{Framework}
\end{figure*}

Fig. \ref{Framework} illustrates a complete framework for drone-assisted monitoring based maintenance planning for wind turbines' blades. However, this article focuses only on the drone-assisted monitoring and online reliability evaluation (highlighted within the dashed box in Fig. \ref{Framework}) parts of the framework. Maintenance planning by considering different factors is out of scope of this article. The steps of the proposed approach are described in the following subsections.

\subsection{Drone Assisted Monitoring}
\label{DroneBasedMonitoring}
As seen in Fig.\ref{Framework}, in the proposed framework, to monitor the wind turbines both on-board monitoring based on traditional approaches and remote monitoring based on UAVs are considered. The on-board condition monitoring systems utilise different sensors embedded into different parts of the turbines to monitor the health of different components and monitoring data are made available to the SCADA system. Data from the SCADA systems can be processed for fault diagnosis and prognosis. As there exist different approaches for such analysis, in this article, we do not cover the analysis of data available in the SCADA system that are received from different sensors. Instead, this article focuses on the processing and analysis of the drone-based monitoring data.
It is assumed that multiple drones will be deployed to monitor the blades of the wind turbines. The drones will inspect the blades and take images of the surface of the blades.
\\
The drone-based blade inspection usually takes 60 to 90 minutes and it should cover four areas of each blade's surface including (a) leading edge, (b) suction side, (c) pressure side and (d) trailing edge. For diagnosis of the blades, aspects that can be considered are: I) type of abnormalities such as crack, erosion, fatigue, flaking, etc., II) depth of the abnormalities that is possible when the Laminate exposes, and III) length, width and the distance from root that can be used also to determine the severity of the issue.
The images taken by drones are then wirelessly transferred to the onshore facilities for further processing. Note that, drones will communicate with each other for a coordinated inspection of the blades and an optimised transmission of data. 

\subsection{Image Processing, Explainability and Confidence Measure}
\label{sec4.2}
After receiving the blade surface images from the drones via a secure channel, the images are processed to identify the anomalies (e.g. fatigue, cracks) on the surface of the blades. There exist several image processing approaches such as \cite{wang2017automatic,peng2018detection,shihavuddin2019wind}  for identifying and characterising the anomalies on the blades. 
Note that, one can use an existing approach or can even develop a new approach. 

\begin{figure*}[bhtb]
      \centering
			\includegraphics[scale=0.4]{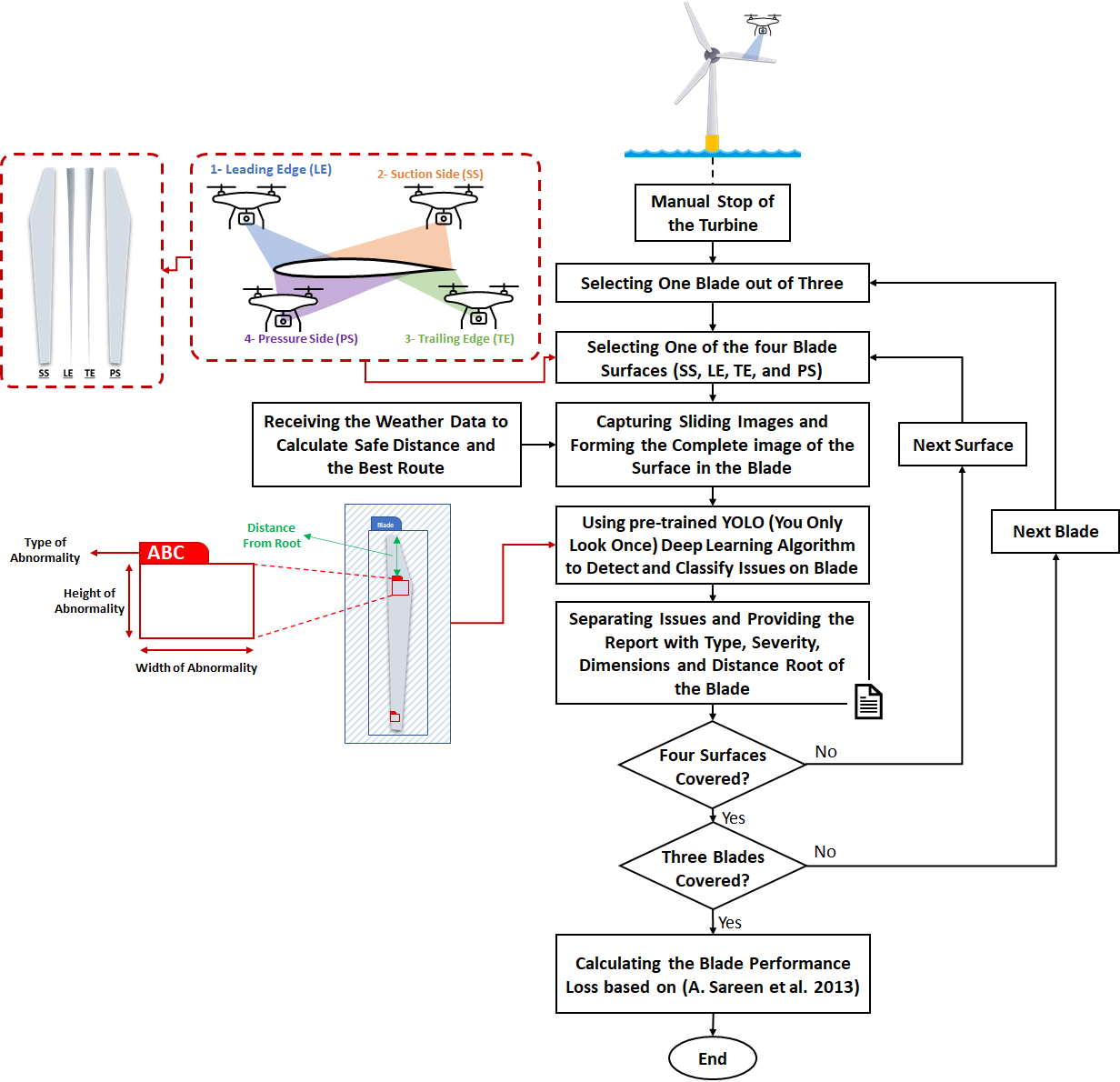}
     \caption{Image processing flowchart for anomaly detection}
      \label{ImageProcessing}
\end{figure*} 

Fig. \ref{ImageProcessing} illustrates the flowchart of the anomaly detection in a wind turbine blade. At the current stage of technology, the turbine needs to be stopped for any type of blade inspection as well as UAV-based inspection. However, having more advanced dynamic positioning may enable UAVs to inspect the blade when the turbine is operational. For each blade, the UAVs will take several pictures of (a) leading edge, (b) suction side, (c) pressure side and (d) trailing edge as mentioned before. The procedure would be similar to the procedure of taking a panorama picture with a mobile phone. In this procedure, the UAV should receive some weather data like wind speed and wind direction to adjust the position, calculate the safe distance from the turbine and avoid the collision. Once all images of a surface is taken, an integrated image of that surface of the blade will be constructed. There are different existing algorithms to process an image and find objects or abnormalities inside the image. For instance, YOLO (You Only Look ONCE) is one of the state-of-the-art algorithms for object detection inside an image with high performance (the performance also depends on the quality of the dataset that the algorithm is trained with) \cite{TrainYourOwnYOLO,redmon2016you}. In the next phase, more information such as type, height, width, distance from root and severity of the abnormalities will be provided for each surface of each blade and the process will be continued until all blades and their surfaces are covered. As an example, for erosion and based on \cite{sareen2014effects} it can be possible to calculate the performance loss of the turbine based on dimensions and severity of detected erosion points on the blade.

When the drones are employed to take images, the quality of the images can be affected by several factors. For instance, images can be blurred due to the relative motion between the blades and the camera mounted on the drones. Noise can be introduced to the images due to the harsh operating conditions of drones and noise produced by various surrounding electronic devices. As a result, the decision made based on these images could be affected depending on the quality of the images. For this reason, we have proposed a way to generate confidence in the decision.

\begin{figure*}[!thpb]
      \centering
			\includegraphics[scale=0.33]{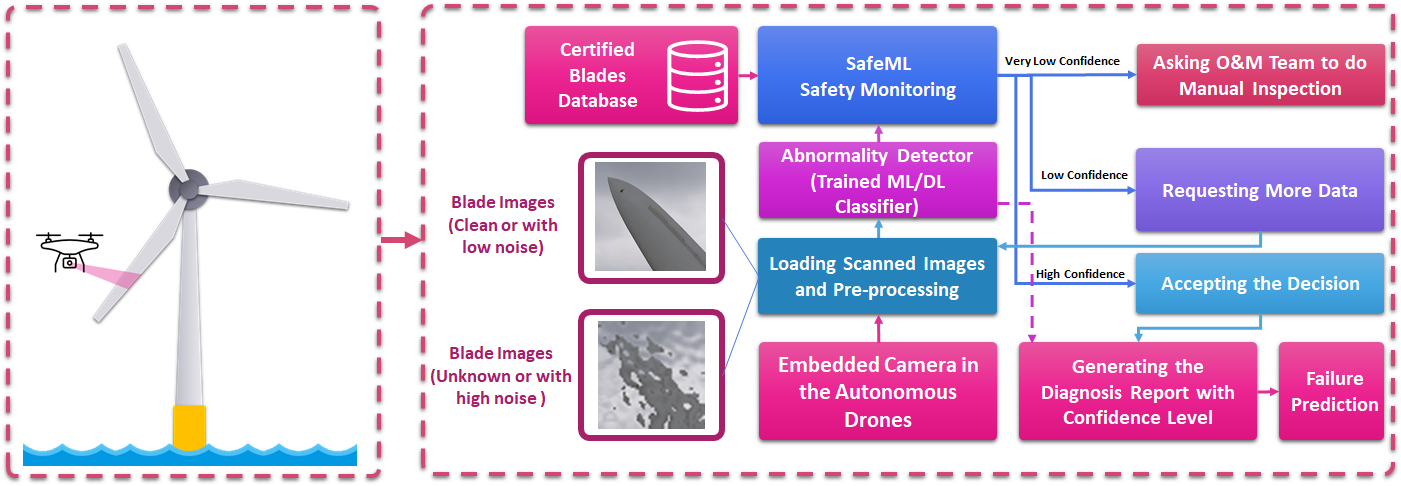}
     \caption{Failure prediction and confidence evaluation}
      \label{ConfidenceEvaluation}
\end{figure*}

Fig. \ref{ConfidenceEvaluation} illustrates the proposed procedure. In this procedure, the images taken by each drone will be loaded into the pre-processing unit and then the pre-processed data will be used as the input of the deep learning algorithm as explained earlier. In the next phase, the SafeML tool (a novel open-source safety monitoring tool \cite{aslansefat2020safeml}) is used to measure the statistical difference between new images and the trusted datasets (the datasets that the deep learning model has been trained with and validated by an expert in the design time) to generate the confidence. Having generated the confidence, three scenarios have been considered; (a) if the confidence is very low, then the approach will provide notice for O\&M team to do the manual inspection, (b) if the confidence is low, the approach will ask the drone to take more pictures from that specific area, and (c) if the confidence is high, the approach will generate the diagnosis report with addressing the evaluated confidence. In the last scenario, the system will be permitted to proceed with the results autonomously. Note that the threshold for the confidence should be tuned in the design time by an expert. The approach is also capable of providing deep learning explainability and interpretability. Due to brevity, we do not provide a detailed description of this capability. 

\subsection{Online Reliability Evaluation}
\label{OnlineReliability}
In this phase, it is assumed that the failure behaviour of the blade system is analysed at design time and reliability model(s) are created. A reliability model could be in the form of a fault tree model or a Markov model or a Bayesian network model or any other existing models. At design time, statistical failure probabilities of different events are used for reliability analysis. Now, for online reliability analysis, we utilise the observations provided by the drone-assisted monitoring. At first, we identify which events or parameters within the existing reliability models are observable by the drones. After that, evidence or observations associated with these events or parameters are extracted from the processing of the data shared by the drones. This online monitoring information about the observable events is then provided as updated inputs to the reliability models to get the updated reliability. In this article, we assumed that an event can either be observed to be in some discrete states with certainty or can be observed to be in different states with different probabilities. For instance, at a certain point in time, a component \textit{A} of a system can be observed to be completely operational or complete failed, i.e., binary observation. On the other hand, the same component can be observed to be said that its failure probability has changed from 0.3 to 0.45, i.e. an increase in prior probability.

\section{Illustrative Example}
\label{sec5}
To demonstrate the proposed approach, we considered the reliability model for blade system failure of offshore wind turbine presented in \cite{liu2019reliability}. In \cite{liu2019reliability}, a fault tree model for the blade system failure was presented. In the fault tree model, 16 different basic events (lowest-level causes) are combined using Boolean AND and OR gates to show the causes of the top event \textit{`blade system failure'}. The basic events and their probabilities 
are shown in Table \ref{BETable}. Note that the occurrence probabilities of the basic events shown in the table are hypothetical and used only for illustrative purposes. However, as the proposed approach is generally applicable to any types of system, when the real operational data from a practical system is available they can be used in the same fashion.

\begin{figure*}[!thpb]
      \centering
			\includegraphics[scale=0.6]{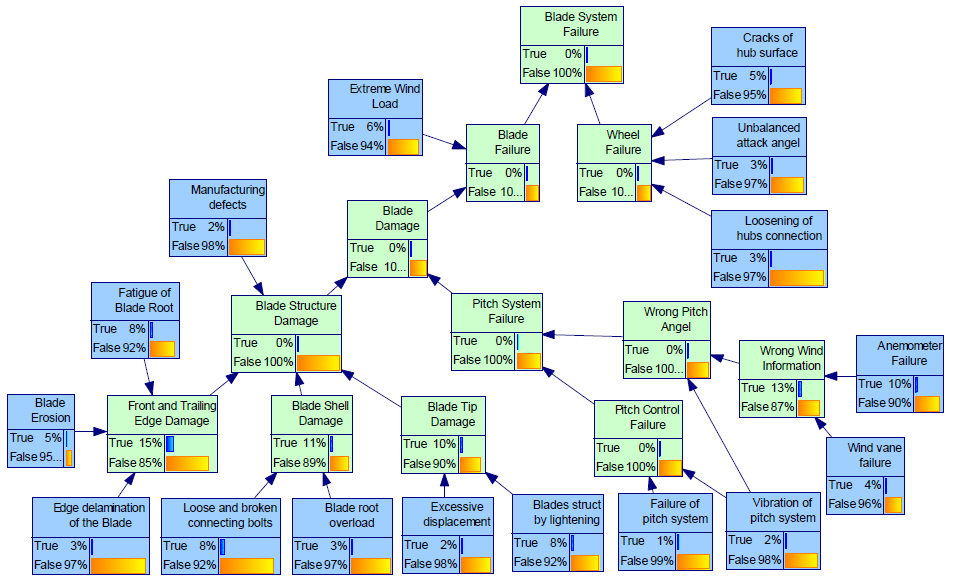}
     \caption{Bayesian network model of the failure of blade system}
      \label{BNModel}
\end{figure*}

For online reliability analysis of the blade system of a wind turbine, in this article, we have mapped the fault tree of \cite{liu2019reliability} into a BN shown in Fig.\ref{BNModel} following the process described in \cite{kabir2018dynamic}. In this model, the blue coloured nodes are the basic events and their prior probabilities are defined based on the probability values reported in Table \ref{BETable}. Note that the basic events are considered as binary, i.e., they can either be in \textit{`True'} (occurred) or \textit{`False'} (non-occurred) state. On the other hand, the green coloured nodes are intermediate events, which are different logic gates in the fault tree. Therefore, the conditional probability tables of these nodes are generated based on the logical behaviour of the logic gates they represent.    

Now, for online reliability analysis, we assume that the events BE1: fatigue of blade root, BE2: blade erosion, and BE14: the cracks of hub surface are observable by the drones, and states of other events are not observable by the drones. That means drone-based monitoring can be utilised to provide evidence about the states of these observable events. Based on the data shown in Table \ref{BETable}, the probability of the blade system failure is obtained as 2.114E-4. As mentioned in section \ref{OnlineReliability}, the proposed approach can consider both deterministic and probabilistic observations about events. To illustrate how deterministic observations can be used to evaluate the online reliability of the blade system, in Table \ref{Result1}, all possible combinations of deterministic observations (8 cases: C1 to C8) regarding BE1, BE2, and BE14 are listed.  Each of these cases is considered one at a time and the evidence are provided in the BN model. Based on this evidence the failure probabilities of the blade system in different scenarios are evaluated and shown in Table \ref{Result1}. The last column of the table shows the \% change of the online probability with compared to the probability evaluated offline, i.e., 2.114E-4. As can be seen in the Table, the probability of the blade system increases sharply when the BE14 is observed to be `True' irrespective of other observations. In other words, when the drones observe cracks on hub surface the failure probability of the blade system increases the most. On the contrary, observing the event BE14 to be `False' leads to a decrease in failure probability irrespective of other observations for the other two events.   

\begin{table}[bhtb]
\renewcommand{\arraystretch}{1.0}
\centering
\caption{Root causes of blade system failure\cite{liu2019reliability} and their  assumed probabilities}
\begin{tabular}{|c| c| c|c|c|c|}
\hline
\textbf{\makecell{Event\\ Tags}}&\textbf{Name}&\textbf{Probability}&\textbf{\makecell{Event\\ Tags}}&\textbf{Name}&\textbf{Probability}\\
\hline
BE1 & Fatigue of blade root& 0.0830 & BE9 & \makecell{The vibration of \\pitch system} & 0.0196\\\hline
BE2 & Blade erosion & 0.0458 & BE10 & \makecell{The failure of \\pitch system} & 0.0116\\\hline
BE3 & \makecell{Edge delamination \\of the blade} & 0.0324 & BE11 & Wind vane failure & 0.0425\\\hline
BE4 & Manufacturing defects & 0.0219 & BE12 & Anemometer failure & 0.0965\\\hline
BE5 & \makecell{Loose and broken\\ of connecting bolts} & 0.0785 & BE13 &  Extreme wind load & 0.0590\\\hline
BE6 & Blade root overload & 0.0296 & BE14 & \makecell{The cracks of \\hub surface} & 0.0466\\\hline 
BE7 & Excessive displacement & 0.0193 & BE15 & \makecell{The unbalanced \\attack angel} & 0.0294\\\hline 
BE8 &  \makecell{The blades struck \\by lightning} & 0.0787 & BE16 & \makecell{Loosening of \\hubs connection} & 0.0305\\\hline 
\end{tabular}
\label{BETable}
\end{table}

\begin{table}[!thpb]
\renewcommand{\arraystretch}{1.5}
\centering
\caption{Online reliability of the blade system in different observed scenarios where observations for events' probabilities are considered as binary}
\begin{tabular}{|l| l| l| l| l| l| l| l|l| l| l| l|}
\hline
\multirow{2}{*}{\textbf{No.}}&\multicolumn{3}{c|}{\textbf{Basic events}}&\multirow{2}{*}{\textbf{BSFP}}& \multirow{2}{*}{\makecell[l]{\textbf{\%}\\\textbf{Change}\\ $(\uparrow \downarrow)$ }}&\multirow{2}{*}{\textbf{No.}}&\multicolumn{3}{c|}{\textbf{Basic events}}&\multirow{2}{*}{\textbf{BSFP}}& \multirow{2}{*}{\makecell[l]{\textbf{\%}\\\textbf{Change}\\ $(\uparrow \downarrow)$ }}\\
\cline{2-4}\cline{8-10}
&\textbf{BE1}&\textbf{BE2}&\textbf{BE14}&&&&\textbf{BE1}&\textbf{BE2}&\textbf{BE14}&&\\
\hline
\textbf{C1} & F & F & F & 1.680E-4 & 20.53\% $\downarrow$&\textbf{C5} & T & F & F & 1.808E-4 & 14.47\% $\downarrow$\\
\hline
\textbf{C2} & F & F & T & 1.065E-3 & 403.78\% $\uparrow$ & \textbf{C6} & T & F & T & 1.077E-3 & 409.46\% $\uparrow$\\
\hline
\textbf{C3} & F & T & F & 1.808E-4 & 14.47\% $\downarrow$ & \textbf{C7} & T & T & F & 1.808E-4 & 14.47\% $\downarrow$\\
\hline
\textbf{C4} & F & T & T & 1.077E-3 & 409.46\% $\uparrow$ & \textbf{C8} & T & T & T & 1.077E-3 & 409.46\% $\uparrow$\\
\hline
\end{tabular}
\begin{tablenotes}
    \item[1] *T:True, F:False, BSFP: Blade system failure probability, $\uparrow:$ Increase, $\downarrow:$ decrease. 
    \end{tablenotes}
\label{Result1}
\end{table}

\begin{figure*}[!thpb]
      \centering
			\includegraphics[scale=0.55]{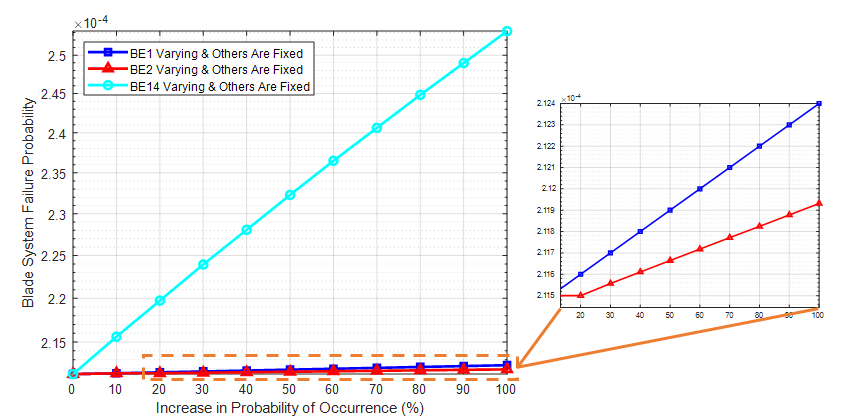}
     \caption{Failure probability of the blade systems with varying levels of probabilities of events BE1, BE2, and BE14 }
      \label{Reliability}
\end{figure*}

Table \ref{Result2} (non-grey cells) shows the results for probabilistic observations. In this case, it was assumed that based on the drone-based monitoring, observations about the changes in occurrence probability of the events can be provided. The change in probability for an event is denoted by $P_i^j$, which means due to the degradation of the components the occurrence probability of an event $i$ is increased by $j\%$ from its prior probability. We performed experiments with many different combinations of such observations for the three observable events and evaluated the failure probability accordingly. Due to space limitation, in Table \ref{Result2}, only 10 randomly selected cases are reported, where online failure probabilities of the blade system are determined for different levels of increase in failure probability of events BE1, BE2, and BE14. At the same time, to evaluate the individual effect of changes of the failure rate of these events, we performed some experiments by changing the probability of one event at a time while keeping other events' probability unchanged. Result of these experiments is shown in Fig. \ref{Reliability}. This figure shows the changes in the online blade system failure probability for a different level of changes to the failure probability of the events BE1, BE2, and BE14, respectively.    

\begin{table}[!thpb]
\renewcommand{\arraystretch}{1.5}
\centering
\caption{Online reliability of the blade system in different observed scenarios where observations for events' probabilities are considered as \% increase from original probabilities presented in Table \ref{BETable} and combination of binary and probability values }
\begin{tabular}{|l| l| l| l| l| l|l| l| l| l| l| l|}
\hline
\multirow{2}{*}{\textbf{No.}}&\multicolumn{3}{c|}{\textbf{Basic events}}&\multirow{2}{*}{\textbf{BSFP}}&\multirow{2}{*}{\makecell[l]{\textbf{\%} \\\textbf{Change}\\ $(\uparrow \downarrow)$ }}&\multirow{2}{*}{\textbf{No.}}&\multicolumn{3}{c|}{\textbf{Basic events}}&\multirow{2}{*}{\textbf{BSFP}}&\multirow{2}{*}{\makecell[l]{\textbf{\%} \\\textbf{Change}\\ $(\uparrow \downarrow)$ }}\\
\cline{2-4}\cline{8-10}
&\textbf{BE1}&\textbf{BE2}&\textbf{BE14}&&&&\textbf{BE1}&\textbf{BE2}&\textbf{BE14}&&\\
\hline
\textbf{C1} & $P_1^5$ & $P_2^5$ & $P_{14}^5$ & 2.135E-4&0.99\% $\uparrow$& \cellcolor{mygrey}\textbf{C11} & \cellcolor{mygrey}$P_1^{10}$ & \cellcolor{mygrey}T &\cellcolor{mygrey} $P_{14}^{15}$ & \cellcolor{mygrey}2.288E-4& \cellcolor{mygrey}8.23\% $\uparrow$\\
\hline
\textbf{C2} &$P_1^{15}$ & $P_2^{15}$ & $P_{14}^{15}$ & 2.179E-4&3.07\% $\uparrow$&\cellcolor{mygrey}\textbf{C12} & \cellcolor{mygrey}$P_1^{30}$ & \cellcolor{mygrey}T & \cellcolor{mygrey}T & \cellcolor{mygrey}1.077E-3& \cellcolor{mygrey}409.46\% $\uparrow$ \\
\hline
\textbf{C3} & $P_1^{25}$ & $P_2^{25}$ & $P_{14}^{25}$ & 2.223E-4&5.16\% $\uparrow$&\cellcolor{mygrey}\textbf{C13} & \cellcolor{mygrey}F & \cellcolor{mygrey}$P_2^{30}$ & \cellcolor{mygrey}F & \cellcolor{mygrey}1.688E-4 & \cellcolor{mygrey}20.15\% $\downarrow$ \\
\hline
\textbf{C4} & $P_1^{50}$ & $P_2^{50}$ & $P_{14}^{50}$ & 2.330E-4&10.22\% $\uparrow$&\cellcolor{mygrey}\textbf{C14} &\cellcolor{mygrey} $P_1^{20}$ &\cellcolor{mygrey} $P_1^{50}$ &\cellcolor{mygrey} T & \cellcolor{mygrey}1.067E-3&\cellcolor{mygrey} 404.73\% $\uparrow$ \\
\hline
\textbf{C5} & $P_1^{75}$ & $P_2^{75}$ & $P_{14}^{75}$ & 2.439E-4&15.37\% $\uparrow$&\cellcolor{mygrey}\textbf{C15} & \cellcolor{mygrey}T & \cellcolor{mygrey}F & \cellcolor{mygrey}$P_{14}^{25}$ & \cellcolor{mygrey}2.330E-4& \cellcolor{mygrey}10.22\% $\uparrow$ \\
\hline
\textbf{C6} & $P_1^{10}$ & $P_2^{5}$ & $P_{14}^{15}$ & 2.178E-4&3.03\% $\uparrow$&\cellcolor{mygrey}\textbf{C16} & \cellcolor{mygrey}T & \cellcolor{mygrey}$P_{2}^{15}$ & \cellcolor{mygrey}$P_{14}^{40}$ &\cellcolor{mygrey} 2.392E-4 &\cellcolor{mygrey} 13.15\% $\uparrow$ \\
\hline
\textbf{C7} & $P_1^{20}$ & $P_2^{10}$ & $P_{14}^{5}$ & 2.137E-4&1.09\% $\uparrow$&\cellcolor{mygrey}\textbf{C17} &\cellcolor{mygrey} $P_1^{30}$ & \cellcolor{mygrey}$P_{2}^{25}$ & \cellcolor{mygrey}F & \cellcolor{mygrey}1.701E-4 & \cellcolor{mygrey}19.54\% $\downarrow$ \\
\hline
\textbf{C8} & $P_1^{15}$ & $P_2^{25}$ & $P_{14}^{15}$ & 2.178E-4&3.03\% $\uparrow$&\cellcolor{mygrey}\textbf{C18} & \cellcolor{mygrey}F & \cellcolor{mygrey}$P_{2}^{25}$ & \cellcolor{mygrey}$P_{14}^{10}$ & \cellcolor{mygrey}2.147E-4 &\cellcolor{mygrey} 1.56\% $\uparrow$ \\
\hline
\textbf{C9} & $P_1^{30}$  &  $P_2^{40}$  & $P_{14}^{25}$ &2.224E-4&5.20\% $\uparrow$&\cellcolor{mygrey}\textbf{C19} &\cellcolor{mygrey} T &\cellcolor{mygrey} T & \cellcolor{mygrey}$P_{14}^{20}$ & \cellcolor{mygrey}2.309E-4 & \cellcolor{mygrey}9.22\% $\uparrow$ \\
\hline
\textbf{C10} & $P_1^{75}$ & $P_2^{20}$  & $P_{14}^{50}$ & 2.331E-4&10.26\% $\uparrow$ &\cellcolor{mygrey}\textbf{C20} & \cellcolor{mygrey} $P_1^{50}$ & \cellcolor{mygrey}F &\cellcolor{mygrey} T & \cellcolor{mygrey}1.066E-3 & \cellcolor{mygrey}404.26\% $\uparrow$\\
\hline
\end{tabular}
\begin{tablenotes}
    \item[1] *$P_i^j:$ probability of event $i$ is increased by $j\%$ from its original value reported in Table \ref{BETable}
    \end{tablenotes}
\label{Result2}
\end{table}

Finally, the grey cells of Table \ref{Result2} shows the results of the 10 randomly selected cases of many experiments performed considering the combination of deterministic and probabilistic observations. As can be seen, in each case, for some events deterministic observations are provided while for others probabilistic observations are provided. Also, for each case, the \% change in the online failure probability is reported. Form the above illustrations, it can be said that using the proposed approach it is possible to utilise the drone-assisted monitoring to collect the operational states of the components of wind turbines, which can, in turn, be used in the reliability models to calculate online reliability that changes over time due to the changes in the operational states of the components. This kind of online reliability will accurately capture the real operational status of the wind turbine systems, thus can assist in effective maintenance scheduling to improve the performance of the systems both in terms of energy production efficiency and cost of energy. 


\section{Conclusion}
This paper proposes an approach for online reliability evaluation of wind turbine systems based on drone-assisted monitoring of the components of the wind turbines. In the approach, drones are deployed to monitor the blades of the wind turbines and images taken by the drones are sent to the onshore processing units for further processing. These images are then processed using machine learning algorithms to detect anomalies on the blade surface. The decision from the anomaly detection process is provided as evidence to the pre-existing reliability models to predict the online failure probability of the blade systems. Using the proposed approach it is possible to utilise the advantages of the drone-based remote inspection of offshore wind turbines and use the computational power of modern computing facilities to process complex images to facilitate online reliability evaluation.      

In this current study, due to lack of access to data from a practical wind turbine system, we used illustrative example with hypothetical data to demonstrate the proposed approach. In the future, we have the plan to verify the effectiveness of this approach through an application to a real system. Moreover, currently, the focus of this paper is only limited to the online failure probability prediction of the blade system. Although the complete framework of Fig. \ref{Framework} shows the steps needed for maintenance schedule planning based on monitoring knowledge and other associated factors, a complete description and illustration of those steps are not considered in this paper. In the future, we plan to address these issues. 

\section*{Acknowledgements}
This work was partly supported by the University of Bradford under SURE Grant scheme and by the Secure and Safe Multi-Robot Systems (SESAME) H2020 Project under Grant Agreement 101017258. We would like to thank EDF Energy R$\&$D UK Centre, AURA Innovation Centre and University of Hull for their support.

\bibliographystyle{splncs04}

\end{document}